\documentclass[runningheads]{llncs}

\usepackage[T1]{fontenc}
\usepackage[T5]{fontenc}

\usepackage{pgfplots}
\usepackage{bbding}
\pgfplotsset{compat=1.18}
\usepackage{sansmath}
\usetikzlibrary{patterns}

\usepackage{bbding}
\usepackage{amsmath}
\usepackage{amssymb} 
\usepackage{graphicx}
\usepackage{array}
\usepackage{booktabs} 
\usepackage{bbding}
\usepackage{algorithm}  
\usepackage{algpseudocode}  
\usepackage{hyperref}
\usepackage{multirow}
\usepackage{color}

\begin{document}
\title{ViHERMES: A Graph-Grounded Multihop Question Answering Benchmark and System for Vietnamese Healthcare Regulations}

%

\titlerunning{ViHERMES}
%
\author{Long S. T. Nguyen\inst{\star}\orcidID{0009-0008-7488-4714}\and Quan M. Bui\inst{\star} \and Tin T. Ngo \and Quynh T. N. Vo  \and Dung N. H. Le\and
Tho T. Quan\textsuperscript{(\Envelope)}\orcidID{0000-0003-0467-6254}}
\renewcommand{\thefootnote}{\fnsymbol{footnote}}
\footnotetext[1]{The two authors contributed equally to this work.}

\setcounter{footnote}{0}
\renewcommand{\thefootnote}{\arabic{footnote}}

\authorrunning{Long S. T. Nguyen et al.}
%
\institute{URA Research Group, Faculty of Computer Science and Engineering, Ho Chi Minh City University of Technology (HCMUT), VNU-HCM, Vietnam }
\maketitle              

\vspace{-0.3cm}

\begin{abstract}
Question Answering (QA) over regulatory documents is inherently challenging due to the need for multihop reasoning across legally interdependent texts, a requirement that is particularly pronounced in the healthcare domain where regulations are hierarchically structured and frequently revised through amendments and cross-references. Despite recent progress in retrieval-augmented and graph-based QA methods, systematic evaluation in this setting remains limited, especially for low-resource languages such as Vietnamese, due to the lack of benchmark datasets that explicitly support multihop reasoning over healthcare regulations. In this work, we introduce the Vietnamese Healthcare Regulations–Multihop Reasoning Dataset (ViHERMES), a benchmark designed for multihop QA over Vietnamese healthcare regulatory documents. ViHERMES consists of high-quality question–answer pairs that require reasoning across multiple regulations and capture diverse dependency patterns, including amendment tracing, cross-document comparison, and procedural synthesis. To construct the dataset, we propose a controlled multihop QA generation pipeline based on semantic clustering and graph-inspired data mining, followed by large language model–based generation with structured evidence and reasoning annotations. We further present a graph-aware retrieval framework that models formal legal relations at the level of legal units and supports principled context expansion for legally valid and coherent answers. Experimental results demonstrate that ViHERMES provides a challenging benchmark for evaluating multihop regulatory QA systems and that the proposed graph-aware approach consistently outperforms strong retrieval-based baselines. The ViHERMES dataset and system implementation are publicly available at \url{https://github.com/ura-hcmut/ViHERMES}.

\end{abstract}

\section{Introduction}

Accessing and complying with regulatory documents is a critical requirement in modern societies, particularly in domains where legal correctness directly affects public safety and service quality. As governments increasingly publish regulations in digital form, there is a growing demand for intelligent \textit{Question Answering} (QA) systems that can assist users in understanding and navigating complex legal and administrative texts \cite{MARTINEZGIL2023100552}.

The healthcare domain presents a particularly challenging setting for regulatory QA. Unlike medical articles or clinical guidelines, which are often descriptive and relatively self-contained, healthcare regulations are legal documents with strict administrative structures and dense inter-document dependencies. These documents are typically organized in a hierarchical manner, with clearly defined units such as articles and clauses, and are frequently interconnected through legal relations such as amendments, replacements, supplements, and cross-references. As a result, regulatory QA in healthcare is inherently a multihop problem rather than a single-document comprehension task. In practice, multihop reasoning becomes a fundamental requirement, as correct answers may depend on identifying the latest effective provision, tracing chains of amendments across multiple documents, and combining definitions or procedural rules referenced from separate regulations, rendering single-hop retrieval or isolated passage matching fundamentally insufficient in this setting \cite{lee-etal-2025-koblex,nguyen-etal-2025-small}.

Recent advances in Artificial Intelligence, particularly in Natural Language Processing, have significantly improved the capabilities of QA systems. \textit{Retrieval-Augmented Generation} (RAG) frameworks have become a dominant paradigm by combining neural retrieval models with \textit{Large Language Models} (LLMs) to generate answers grounded in retrieved contexts \cite{10.5555/3495724.3496517}. While naive RAG approaches enhance factual coverage, they typically treat documents as flat collections of passages and lack explicit awareness of legal structures and inter-document relations \cite{10.1145/3777378}. In regulatory domains, this limitation often leads to answers that are linguistically plausible but legally incomplete or outdated. To address these shortcomings, graph-based retrieval and data mining methods such as GraphRAG \cite{edge2024local}, HippoRAG \cite{gutierrez2025from}, and LightRAG \cite{guo-etal-2025-lightrag} have been proposed to incorporate structured knowledge into the retrieval process. However, when applied to regulatory documents, such automatically induced graphs often struggle to faithfully capture the formal and hierarchical nature of legal texts \cite{de2025graph}. In particular, multihop regulatory questions that depend on amendment chains or legal validity require precise, rule-aware relationships that are difficult to recover from generic entity–relation extraction alone.

Vietnamese is a low-resource language, and datasets that jointly address the healthcare and legal or regulatory domains remain particularly scarce. Despite the practical importance of multihop regulatory QA, the lack of high-quality benchmark datasets has hindered systematic research in this area. Existing Vietnamese legal QA datasets do include regulatory texts; however, they primarily focus on general legal documents or regulations in other domains such as education, and are typically designed for single-document comprehension or shallow QA rather than multihop reasoning across documents \cite{11063484,Do2025,11063637,IJAI25255,minh-etal-2024-using}. Similarly, existing Vietnamese healthcare QA datasets largely concentrate on clinical narratives, medical records, or healthcare-related news articles, rather than healthcare regulatory documents \cite{tran-etal-2024-vimedaqa,10.1145/3527631,10.1007/978-3-031-10986-7_30}. Crucially, none of these datasets explicitly support multihop reasoning over legally interdependent healthcare regulations. As a result, the performance and analytical understanding of QA systems in this domain remain difficult to evaluate in a controlled and meaningful manner.

In this work, we introduce \textit{\textbf{Vi}etnamese \textbf{HE}althcare \textbf{R}egulations–\textbf{M}ultihop R\textbf{E}asoning Data\textbf{S}et} (ViHERMES), the first benchmark dataset specifically designed for multihop QA over Vietnamese healthcare regulations. ViHERMES focuses on questions that inherently require reasoning across multiple regulatory documents, capturing diverse dependency patterns such as amendment tracing, cross-document comparison, and procedural synthesis. To construct the dataset, we propose a controlled multihop QA generation pipeline that leverages semantic clustering as a form of graph-inspired data mining to sample coherent sets of regulatory contexts, followed by LLM-based QA generation with structured evidence and reasoning annotations. Beyond the dataset, we also propose a graph-aware retrieval framework tailored to the hierarchical and legally constrained nature of regulatory documents. Our approach represents regulations at the level of legal units and explicitly models formal legal relations, enabling principled context expansion strategies that respect legal validity and avoid context drift. This retrieval mechanism is integrated into a multi-agent QA system that separates intent understanding, graph-based retrieval, and answer verification, thereby supporting robust and intelligent healthcare regulatory QA. Our main contributions are summarized as follows.

\begin{itemize}
  \item We introduce ViHERMES, the first benchmark dataset for Vietnamese healthcare regulatory QA that explicitly targets multihop reasoning across legally interdependent documents.
  \item We propose a graph-inspired data mining pipeline for controlled multihop QA dataset construction with high-quality evidence-grounded annotations.
  \item We present a graph-aware, multi-agent QA framework that effectively leverages legal structure and demonstrates consistent improvements over strong retrieval-based baselines on ViHERMES.
\end{itemize}

\section{Related Works}

\subsection{Multihop QA and Graph-based Retrieval}

Multihop QA has been extensively studied to address queries that require reasoning over multiple pieces of evidence rather than a single passage. Early approaches such as IRCoT \cite{trivedi-etal-2023-interleaving} interleave retrieval with chain-of-thought reasoning to iteratively guide evidence selection, while RAPTOR \cite{sarthi2024raptor} organizes textual units into hierarchical summaries that support retrieval across different levels of abstraction. Building upon these ideas, recent graph-based retrieval frameworks, including GraphRAG \cite{edge2024local}, HippoRAG \cite{gutierrez2025from}, LightRAG \cite{guo-etal-2025-lightrag}, and MiniRAG \cite{fan2025minirag}, incorporate graph structures into retrieval-augmented generation by modeling entities, text chunks, or their relations as nodes and enabling neighborhood expansion or graph-based ranking. These methods have demonstrated strong performance gains on general multihop QA benchmarks and improved efficiency, particularly in resource-constrained settings. However, most existing approaches are primarily designed for unstructured or loosely structured text collections and rely on automatically induced, largely entity-centric graphs. Consequently, they struggle to faithfully capture the formal hierarchy, rule-based dependencies, and legal validity constraints inherent in regulatory documents. This limitation is especially pronounced in the healthcare domain, where answering a question often requires tracing amendment chains, identifying currently effective provisions, and combining legally interdependent clauses across multiple documents. In contrast, our work explicitly aligns both dataset construction and retrieval with the intrinsic legal structure of healthcare regulations, enabling principled multihop reasoning over legally interdependent documents.

\subsection{Vietnamese Regulation-related QA Datasets}

Benchmark datasets for Vietnamese QA remain relatively scarce, particularly for domains involving structured regulatory documents. Existing datasets that cover regulations are primarily situated in the legal or educational domains. ALQAC \cite{11063484} introduces a manually annotated legal QA dataset based on Vietnamese statute laws, while several other works focus on regulations in specific domains, including educational management \cite{minh-etal-2024-using}, bidding law (ViBidLQA) \cite{11063637}, university training regulations (ViRHE4QA) \cite{Do2025}, and large-scale labor law retrieval with associated QA data \cite{IJAI25255}. Although these datasets provide valuable resources for Vietnamese legal QA, they are typically designed for single-document comprehension or shallow question answering and do not explicitly support multihop reasoning across legally interdependent regulations. In contrast, existing Vietnamese healthcare QA datasets largely focus on clinical or informational content rather than regulatory texts. ViMedAQA \cite{tran-etal-2024-vimedaqa} targets abstractive medical QA over clinical topics such as diseases and drugs, UIT-ViNewsQA \cite{10.1145/3527631} is constructed from healthcare news articles for machine reading comprehension, and ViHealthQA \cite{10.1007/978-3-031-10986-7_30} collects expert-answered questions from health websites. While these datasets are valuable for healthcare-related QA, they do not reflect the hierarchical structure, amendment relationships, or cross-document dependencies that characterize healthcare regulations. Consequently, none of the existing Vietnamese datasets jointly address healthcare regulations and multihop reasoning over legally interdependent documents, leaving a critical gap that our ViHERMES dataset is specifically designed to fill.

\section{ViHERMES Dataset}
\label{sec:vihermes}

\subsection{Dataset Creation}
\label{sec:dataset_creation}

\begin{figure}[t]
    \centering
    \includegraphics[width=\linewidth]{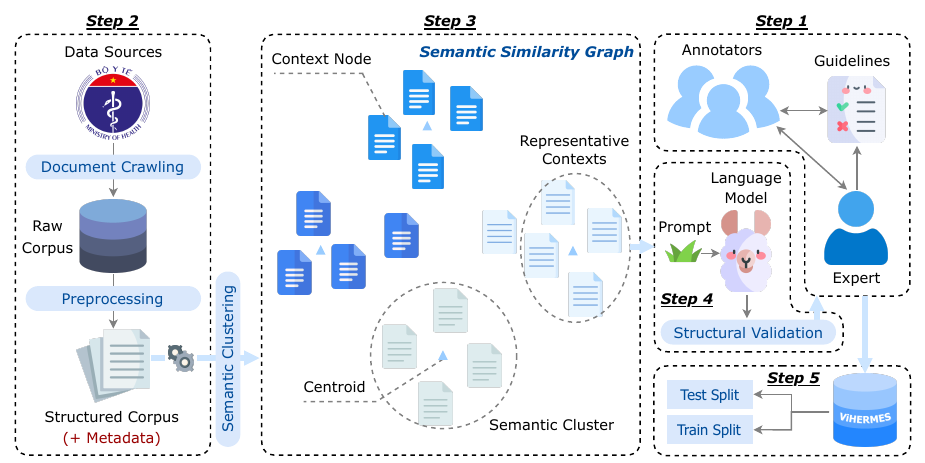}
    \caption{ViHERMES dataset construction pipeline, from corpus collection and semantic clustering to LLM-based generation and splitting.}
    \label{fig:vihermes_pipeline}
\end{figure}

We construct ViHERMES through a controlled dataset creation pipeline that integrates human supervision, semantic clustering, and LLM-based generation to ensure genuine multihop reasoning and evidence-grounded answers over healthcare regulatory documents. An overview of the pipeline is shown in Figure~\ref{fig:vihermes_pipeline}.

\smallskip
\noindent \textbf{Step 1: Annotator recruitment and guideline design.}\quad
We recruit a small team of \textit{annotators} together with a domain \textit{expert}. The expert provides high-level guidance to ensure multihop validity and legal consistency. The annotation process follows shared \textit{guidelines} defining multihop criteria, evidence usage, and legal validity constraints.

\smallskip
\noindent
\textbf{Step 2: Corpus collection and preprocessing.}\quad
Regulatory documents are collected from official \textit{data sources}, such as public portals of the Vietnamese Ministry of Health\footnote{\url{https://moh.gov.vn/web/ministry-of-health}}, via automated \textit{document crawling}. The collected texts form a \textit{raw corpus}, which is subsequently processed by a \textit{preprocessing} pipeline that normalizes content, removes noise, and segments documents according to their legal hierarchy. The resulting \textit{structured corpus}, together with associated \textit{metadata}, consists of context units corresponding to legal units (e.g., articles or clauses), each represented as a context triple $\langle \texttt{id}, \texttt{title}, \texttt{text} \rangle$.

\smallskip
\noindent
\textbf{Step 3: Semantic clustering and similarity graph induction.}\quad
Each context is treated as a \textit{context node} and encoded into an embedding vector $\mathbf{e}_i \in \mathbb{R}^d$ using a neural embedding model. To emphasize semantic direction rather than vector magnitude, embeddings are $\ell_2$-normalized as $\tilde{\mathbf{e}}_i = \mathbf{e}_i / \|\mathbf{e}_i\|_2$. We then perform \textit{semantic clustering} using $K$-Means on the normalized vectors, which corresponds to \textit{Spherical $K$-Means} in practice. Each resulting \textit{semantic cluster} $C_k$ is summarized by its \textit{centroid} $\mathbf{c}_k = \frac{1}{|C_k|} \sum_{\tilde{\mathbf{e}}_i \in C_k} \tilde{\mathbf{e}}_i$. Contexts are ranked by their Euclidean distance to the centroid $d(\tilde{\mathbf{e}}_i, \mathbf{c}_k) = \|\tilde{\mathbf{e}}_i - \mathbf{c}_k\|_2$, and the nearest ones are selected as \textit{representative contexts}. This cluster-and-centroid organization induces an implicit \textit{semantic similarity graph} that preserves topical coherence among regulatory texts. The graph is not an explicit legal graph; rather, it provides a latent similarity structure without relying on potentially noisy automatic extraction of legal relations.

\smallskip
\noindent
\textbf{Step 4: Multihop QA generation and structural validation.}\quad
For a given hop level $h$, we sample tuples of contexts $\mathcal{T} = \{c_1, \dots, c_h\}$ from the pool of \textit{representative contexts} under two constraints. First, contexts within a tuple must originate from different document \texttt{titles}, i.e., $\forall i \neq j,\ \texttt{title}(c_i) \neq \texttt{title}(c_j)$, to ensure cross-document reasoning. Second, to maintain diversity, the overlap between any two tuples $\mathcal{T}_a$ and $\mathcal{T}_b$ is bounded by $|\mathcal{T}_a \cap \mathcal{T}_b| / |\mathcal{T}_a| \le \tau$. Each tuple is then provided to a \textit{language model} through a structured \textit{prompt} to generate multihop question--answer pairs, where each answer must rely on all $h$ contexts. In practice, each hop corresponds to one legal dependency step, such as amendment tracing, definition lookup, or cross-document procedural composition. The generated outputs undergo \textit{structural validation}, which verifies format correctness, consistency between hop level and evidence usage, and explicit reasoning. Ambiguous or borderline cases are further reviewed by annotators under expert supervision to ensure legal coherence and correctness.

\smallskip
\noindent
\textbf{Step 5: Dataset aggregation and splitting.}\quad
After validation, the accepted question--answer instances are aggregated and exported in a structured format. The dataset is then divided into a \textit{train split} and a \textit{test split}, with controlled distributions over hop levels, forming the final \textit{ViHERMES} benchmark.

\subsection{Dataset Statistics}
\label{sec:dataset_statistics}

Table~\ref{tab:vihermes_example} presents a representative instance from the ViHERMES dataset, including the query, answer, number of reasoning hops, explicit reasoning, supporting evidence, and corresponding context identifiers.
Each identifier maps to a regulatory context stored as a triple $\langle \texttt{id}, \texttt{title}, \texttt{text} \rangle$. The \texttt{reasoning} and \texttt{evidence} fields provide explicit explanations and extractive textual spans grounded in the referenced contexts, respectively. Both fields are generated during dataset construction (Step~4) and subsequently verified by annotators, improving transparency and answer reliability. Table~\ref{tab:vihermes_stats} summarizes key statistics of the ViHERMES test set across different \textit{Numbers of Hops} (NoH). For each hop level, we report the number of samples, vocabulary size, and length statistics of questions, answers, and supporting contexts.
As NoH increases, question and answer lengths grow substantially, reflecting higher reasoning complexity, while context lengths expand rapidly, indicating the need to integrate information from long and diverse regulatory texts.

\begin{table*}[t]
\centering
\caption{A representative multihop QA instance from the ViHERMES dataset.}
\resizebox{\textwidth}{!}{
\begin{tabular}{p{2.5cm} p{13.8cm}}
\toprule
\textbf{Field} & \textbf{Content} \\
\midrule
Query &
Để đảm bảo tuân thủ pháp luật một cách toàn diện, một cơ sở khám bệnh, chữa bệnh cần thực hiện những nghĩa vụ chính nào liên quan đến việc áp dụng giá dịch vụ trong giai đoạn chuyển tiếp, việc duy trì chất lượng cơ bản, và các điều kiện pháp lý để được phép hoạt động? \\
& \textbf{English:} To ensure comprehensive legal compliance, what main obligations must a healthcare facility fulfill regarding service pricing during the transition period, maintenance of baseline quality, and the legal conditions required for lawful operation? \\
\midrule
Answer &
Để tuân thủ pháp luật toàn diện, một cơ sở khám bệnh, chữa bệnh phải thực hiện các nghĩa vụ chính sau:
(1) Về điều kiện hoạt động: Phải được cấp giấy phép hoạt động và bảo đảm các y, bác sĩ làm việc tại cơ sở có chứng chỉ hành nghề hợp lệ.
(2) Về chất lượng cơ bản: Phải triển khai các quy định về tiêu chuẩn chất lượng, thực hiện rà soát, khắc phục để duy trì chất lượng, và tiến hành tự đánh giá, báo cáo theo quy định.
(3) Về giá dịch vụ trong giai đoạn chuyển tiếp: Khi có quyết định giá mới, đối với người bệnh đang điều trị từ trước đó, cơ sở phải tiếp tục áp dụng mức giá cũ cho đến khi người bệnh ra viện hoặc kết thúc đợt điều trị ngoại trú. \\
& \textbf{English:} To fully comply with legal requirements, a healthcare facility must fulfill the following obligations:
(1) Operational conditions: the facility must obtain an operating license, and all practicing physicians must hold valid professional certificates.
(2) Baseline quality: the facility must implement quality standards, conduct reviews and corrective actions, and perform self-assessments with mandatory reporting as required.
(3) Service pricing during the transition period: when a new pricing decision is issued, patients who began treatment earlier must continue to be charged the previous price until discharge or completion of outpatient treatment. \\
\midrule
Number of hops & 3 \\
\midrule
Context IDs & [4814-7506-4302, 7115-6545-3240, 6442-4111-1890] \\
\midrule
Evidence &
\texttt{context\_1:} Quy định về áp dụng giá trong giai đoạn chuyển tiếp: ``Tiếp tục được áp dụng mức giá dịch vụ khám bệnh, chữa bệnh theo quy định của cấp có thẩm quyền trước thời điểm thực hiện mức giá mới cho đến khi ra viện hoặc kết thúc đợt điều trị ngoại trú.'' \\
& \texttt{context\_2:} Trách nhiệm của cơ sở khám bệnh, chữa bệnh về chất lượng: ``Triển khai thực hiện quy định về tiêu chuẩn chất lượng; rà soát, bổ sung và khắc phục để duy trì chất lượng khám bệnh, chữa bệnh ở mức cơ bản; thực hiện đánh giá và báo cáo kết quả theo quy định.'' \\
& \texttt{context\_3:} Yêu cầu về điều kiện pháp lý để hoạt động: ``Cơ sở phải được cấp phép hoạt động; các y, bác sĩ thực hiện khám bệnh, chữa bệnh phải có chứng chỉ hành nghề theo quy định.'' \\
& \textbf{English:} The evidence consists of three regulatory provisions covering service pricing during transitional periods, baseline quality responsibilities, and legal requirements for operation, respectively. \\
\midrule
Reasoning &
Câu trả lời tổng hợp các nghĩa vụ của một cơ sở khám bệnh, chữa bệnh từ ba văn bản khác nhau. \texttt{context\_3} cung cấp các điều kiện pháp lý tiên quyết để được phép hoạt động. \texttt{context\_2} xác định các trách nhiệm liên quan đến việc duy trì và báo cáo chất lượng cơ bản. \texttt{context\_1} đưa ra quy định thủ tục về việc áp dụng giá dịch vụ trong giai đoạn chuyển tiếp. Việc kết hợp cả ba nguồn là cần thiết để hình thành một câu trả lời đầy đủ và nhất quán về mặt pháp lý. \\
& \textbf{English:} The answer integrates three regulatory sources covering operational legality, baseline quality, and transitional service pricing, forming a legally consistent response. \\
\bottomrule
\end{tabular}
}
\label{tab:vihermes_example}
\end{table*}

\setlength{\tabcolsep}{5pt}

\begin{table*}[t]
\centering
\caption{Statistics of the ViHERMES test split across different multihop settings. Lengths are reported as \textit{min / mean / max}.}
\label{tab:vihermes_stats}
\resizebox{\textwidth}{!}{
\begin{tabular}{cccccc}
\toprule
\textbf{NoH} & \textbf{NoS} & \textbf{Vocab} & \textbf{Question length} & \textbf{Answer length} & \textbf{Context length} \\
\midrule
1 & 312 & 2232 & 48 / 113.2 / 215 & 58 / 181.8 / 504 & 195 / 1185.1 / 4257 \\
2 & 312 & 2949 & 120 / 249.4 / 446 & 89 / 399.1 / 810 & 734 / 2729.4 / 10075 \\
3 & 312 & 3178 & 168 / 325.7 / 688 & 135 / 524.0 / 1100 & 743 / 3739.8 / 14767 \\
4 & 312 & 3545 & 179 / 386.4 / 813 & 175 / 748.0 / 1696 & 1461 / 5677.5 / 16556 \\
5 & 312 & 3707 & 194 / 445.9 / 977 & 101 / 889.9 / 1938 & 1115 / 6451.6 / 18878 \\
\midrule
All & 1560 & 5703 & 48 / 304.2 / 977 & 58 / 548.8 / 1938 & 195 / 3953.9 / 18878 \\
\bottomrule
\end{tabular}}
\end{table*}

\section{Graph-Aware System for Regulatory QA}
\label{sec:system}

\begin{figure}[t]
\centering
\includegraphics[width=\linewidth]{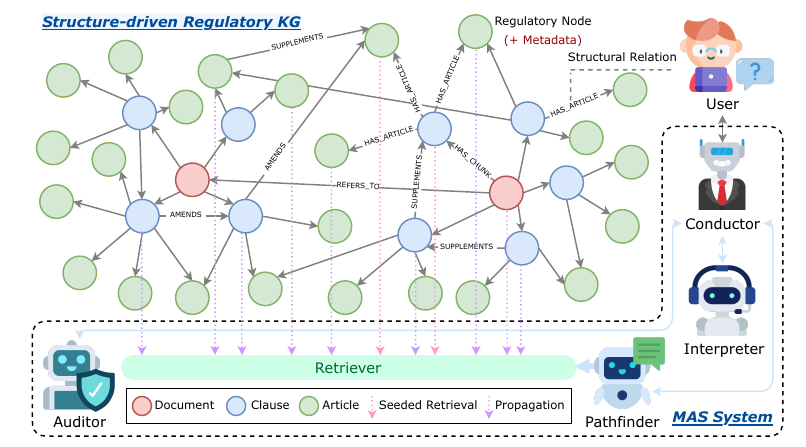}
\caption{
Overview of the structure-driven regulatory QA system.
}
\label{fig:vihermes_system}
\end{figure}

Figure~\ref{fig:vihermes_system} illustrates the proposed \emph{graph-aware QA system} for Vietnamese healthcare regulations. Unlike standard RAG pipelines that treat the corpus as a flat collection of passages, our system operates over a \emph{structure-driven regulatory knowledge graph}, where legal texts are modeled as interconnected \emph{regulatory nodes} linked by defined relations. The system follows a \emph{seeded retrieval and propagation} paradigm, retrieving legally valid and contextually complete evidence.

\subsection{Structure-Driven Regulatory Knowledge Graph}
\label{sec:legal_graph}

The system is built upon a \emph{Structure-driven Regulatory Knowledge Graph} (SRKG), which encodes regulatory documents according to formal legal drafting conventions. Unlike entity-centric or LLM-induced graphs, SRKG is \emph{structure-driven}, with nodes and edges corresponding to legally grounded units and relations. The graph is organized as a \emph{flat regulatory base layer}, where legal-unit nodes coexist on a unified plane and are retrieved directly, rather than through a top-down pipeline. Formally, SRKG is defined as a directed labeled graph $G = (V, E)$, where $V$ denotes regulatory nodes and $E \subseteq V \times \mathcal{R} \times V$ represents typed edges with labels from $\mathcal{R} = \mathcal{R}_{\text{struct}} \cup \mathcal{R}_{\text{legal}}$.

\smallskip
\noindent\textbf{Regulatory nodes.}\quad
SRKG represents regulatory content using \emph{regulatory nodes} augmented with provenance metadata.
As illustrated in Figure~\ref{fig:vihermes_system}, we distinguish three conceptual levels:
\emph{Document nodes} representing legal documents,
\emph{Clause nodes} corresponding to legal units with standalone meaning,
and \emph{Article nodes} capturing finer-grained units where applicable.
Each node $v \in V$ is represented as $v = \langle \texttt{id}(v), \texttt{text}(v), \texttt{meta}(v) \rangle$, where $\texttt{meta}(v)$ includes document identifier, promulgation date, issuing authority, and legal status.
Each node is assigned a deterministic identifier following a strict scheme (e.g., \texttt{DocID::UnitID}), enabling stable referencing, amendment-chain tracing, and incremental ingestion.

\smallskip
\noindent\textbf{Relations.}\quad
SRKG encodes two complementary categories of relations, namely \emph{structural relations} and \emph{legal semantic relations}.
Structural relations $\mathcal{R}_{\text{struct}}=\{\textsc{Has\_Article}, \textsc{Has\_Clause}\}$ preserve legal hierarchy and provenance.
Legal semantic relations $\mathcal{R}_{\text{legal}}=\{\textsc{Amends}, \textsc{Replaces}, \textsc{Supplements}, \textsc{Refers\_To}\}$ capture inter-document dependencies affecting legal interpretation and validity.
All relations are induced using rule-based pattern matching over standardized legal expressions
(e.g., ``sửa đổi Điều~X'' [amends Article~X], ``bổ sung Khoản~Y'' [supplements Clause~Y], ``căn cứ Điều~Z'' [refers to Article~Z]), yielding high precision and avoiding spurious links. When a regulatory node references a legal unit not yet present in the corpus, the system creates a lightweight \emph{placeholder node} to preserve the structural dependency. Such placeholders are automatically resolved when the corresponding documents are later ingested, ensuring graph completeness without disrupting downstream retrieval.

\subsection{Seeded Retrieval and Propagation}
\label{sec:retrieval}

Given a user query $q$, the system performs retrieval through a two-stage
\emph{seeded retrieval and propagation} strategy.

\smallskip
\noindent\textbf{Seeded retrieval.}\quad
Retrieval is first conducted over the legal-unit base layer.
The \emph{Pathfinder} identifies a small set of highly relevant
\emph{seed regulatory nodes}
$S_K(q)=\{v_1,\dots,v_K\} \subseteq V$
by ranking nodes using a hybrid relevance score
$s(v,q)=\lambda\, s_{\text{dense}}(v,q) + (1-\lambda)\, s_{\text{sparse}}(v,q)$,
where $s_{\text{dense}}$ measures semantic similarity in embedding space and
$s_{\text{sparse}}$ captures lexical evidence (e.g., BM25).
The resulting $S_K(q)$ serves as anchor points for downstream expansion.

\smallskip
\noindent\textbf{Propagation.}\quad
Starting from $S_K(q)$, the system expands context in a controlled manner along relation-specific edges.
For a node $v$ and relation $r$, outgoing neighbors are denoted as
$\mathcal{N}_r(v)=\{u \mid (v,r,u)\in E\}$.
This process activates three principled flows:
\begin{itemize}
    \item \textbf{Validity tracing.}\quad
    For $r \in \{\textsc{Amends}, \textsc{Replaces}, \textsc{Supplements}\}$,
    amendment chains are recursively traversed until a terminal node is reached.
    A node $u$ is terminal if $\forall r' \in \{\textsc{Amends}, \textsc{Replaces}, \textsc{Supplements}\},\ \mathcal{N}_{r'}(u)=\emptyset$,
    and such terminal nodes are preferred as legally applicable evidence.
    \item \textbf{Contextual supplementation.}\quad
    For reference relations ($r=\textsc{Refers\_To}$),
    expansion is restricted to direct neighbors only,
    providing essential legal context while preventing uncontrolled drift.
    \item \textbf{Provenance retrieval.}\quad
    Structural relations are used to retrieve upstream document metadata,
    ensuring provenance-aware evidence usage.
\end{itemize}

\smallskip
\noindent\textbf{Context assembly.}\quad
The final retrieval output is assembled as a bounded context set
$C(q)=S_K(q)\cup P(q)$,
where $P(q)$ denotes nodes obtained via relation-aware expansion.
Propagation depth and relation types are explicitly constrained
before $C(q)$ is passed to downstream agents.

\subsection{Multi-Agent System}
\label{sec:multi_agent}

The QA pipeline is implemented as a lightweight \emph{Multi-Agent System} (MAS)
composed of functionally specialized agents, as illustrated in Figure~\ref{fig:vihermes_system}.
Each agent fulfills a distinct operational role within a unified regulatory QA workflow.

\smallskip
\noindent\textbf{Interpreter.}\quad
The \textit{Interpreter} performs intent analysis and routing.
It determines whether a query is regulatory, extracts key signals such as document identifiers,
and decides whether graph-based retrieval is required.
This step avoids unnecessary retrieval, improving overall execution efficiency.

\smallskip
\noindent\textbf{Pathfinder.}\quad
The \textit{Pathfinder} implements the core graph-aware retrieval logic.
It performs seeded retrieval over regulatory nodes followed by relation-aware propagation
on the SRKG,
including validity tracing along amendment relations
and bounded expansion over reference relations.
The agent returns a legally coherent and up-to-date context set $C(q)$.

\smallskip
\noindent\textbf{Auditor.}\quad
The \textit{Auditor} verifies the retrieved evidence and intermediate outputs
by checking consistency between regulatory nodes, relations, and generated claims.
It performs grounding validation and safety checks to detect unsupported
or potentially hallucinated content.

\smallskip
\noindent\textbf{Conductor.}\quad
The \textit{Conductor} orchestrates the overall QA process.
Based on the verified context $C(q)$, it invokes an LLM to generate
the final natural-language answer and coordinates interactions among agents.
When insufficient grounding is detected by the \textit{Auditor},
the \textit{Conductor} enforces conservative behavior such as abstention
or clarification requests.

\smallskip
Overall, the proposed system integrates \emph{structure-driven regulatory representation},
\emph{seeded retrieval}, and \emph{relation-aware propagation} within a unified
multi-agent framework.
By operationalizing legal structure directly during retrieval,
the system enables legally valid multihop reasoning and robust evidence grounding,
as demonstrated in our experiments.

\section{Experimentations}
\label{sec:experiments}

We assess system performance along multiple complementary dimensions,
including answer correctness, multihop reasoning quality, inference latency,
graph construction overhead, and the effectiveness of agent coordination,
and compare against representative baseline approaches.

\subsection{Dataset}
\label{sec:exp_dataset}

All experiments are conducted on the test split of ViHERMES,
which is specifically designed to assess multihop reasoning
over Vietnamese healthcare regulatory documents.
The test set spans diverse hop levels and legal reasoning patterns,
including amendment tracing, cross-document dependency resolution,
and validity-aware clause composition.

\subsection{Baselines}
\label{sec:baselines}

We compare the proposed system against representative baselines
drawn from three major families of retrieval-augmented
and multihop QA approaches.

\smallskip
\noindent\textbf{Naive RAG.}\quad
We consider standard RAG pipelines based on flat passage retrieval,
including (1) lexical retrieval using BM25,
(2) dense retrieval via embedding similarity,
and (3) hybrid retrieval combining lexical and dense scores
through weighted interpolation.
These baselines represent common RAG settings
without explicit multihop or structural awareness.

\smallskip
\noindent\textbf{Reasoning-guided multihop QA.}\quad
IRCoT \cite{trivedi-etal-2023-interleaving} is a multi-step QA approach
that interleaves retrieval with \textit{Chain-of-Thought} (CoT) reasoning
to iteratively guide evidence selection using a strong LLM,
without explicitly modeling document structure or legal validity constraints.

\smallskip
\noindent\textbf{Graph-based RAG methods.}\quad
We include several representative graph-aware retrieval frameworks
that incorporate graph structures into indexing and retrieval
to support multihop reasoning:
\begin{itemize}
    \item \textbf{RAPTOR} \cite{sarthi2024raptor} constructs a hierarchical tree
    of textual units via recursive summarization,
    enabling retrieval across multiple levels of abstraction.

    \item \textbf{MiniRAG} \cite{fan2025minirag} is a lightweight graph-based system
    that combines a small language model with a heterogeneous graph index
    to achieve efficient structured retrieval under limited computational budgets.

    \item \textbf{LightRAG} \cite{guo-etal-2025-lightrag} emphasizes efficient graph
    construction and traversal strategies to balance retrieval quality
    and inference latency.

    \item \textbf{HippoRAG~2} \cite{gutierrez2025from} draws inspiration from
    hippocampal memory mechanisms, modeling long-term knowledge as a graph
    to improve recall and evidence integration across distant reasoning hops.
\end{itemize}

\subsection{Evaluation Metrics}
\label{sec:metrics}

We evaluate system performance at both the answer and retrieval levels.
For answer quality, we report token-level $F1$ to measure surface-level overlap
between predicted and reference answers.
Since $F1$ alone is insufficient to assess reasoning correctness and legal validity
in multihop regulatory QA, we additionally adopt an \emph{LLM-as-a-Judge}
evaluation protocol~\cite{li-etal-2025-generation}, where a strong external LLM
evaluates answers in terms of correctness, completeness, and consistency with
supporting evidence.
To assess retrieval quality independently of answer generation, we use
$\mathrm{Recall@}k = \frac{|E_{\text{gold}} \cap E_{\text{retrieved}}^{(k)}|}{|E_{\text{gold}}|}$,
which measures the proportion of gold supporting contexts recovered within the
top-$k$ retrieved results and reflects the effectiveness of evidence retrieval
for multihop reasoning.

\subsection{Experimental Setup}
\label{sec:exp_setup}

To ensure fair comparison and reproducibility,
all systems share the same language model backbone
and evaluation protocol.
We adopt \texttt{GPT-4o-mini}\footnote{\url{https://platform.openai.com/docs/models/gpt-4o-mini}}
as the unified LLM backbone for answer generation,
chosen for its balance between reasoning capability
and computational efficiency.
Text embeddings are generated using
OpenAI \texttt{text-embedding-3-small}\footnote{\url{https://platform.openai.com/docs/models/text-embedding-3-small}},
a multilingual embedding model supporting both Vietnamese and English.
All baselines are evaluated using default hyperparameters
to reflect realistic out-of-the-box performance.
LLM-as-a-Judge evaluations are conducted using the
\texttt{GPT-4o}\footnote{\url{https://platform.openai.com/docs/models/gpt-4o}} API
under a standardized rubric to ensure consistent
and comparable scoring across systems.

\setlength{\tabcolsep}{3pt}

\begin{table}[t]
\centering
\caption{Main QA performance on the ViHERMES test set.}
\label{tab:main_results}
\begin{tabular}{lccc}
\toprule
\textbf{Method} & $\mathbf{F1}$ & \textbf{LLM Judge} & $\mathbf{Recall@5}$ \\
\midrule
Naive RAG (BM25)   & 0.3076 & 0.2027 & 0.2617 \\
Naive RAG (Dense)  & 0.3289 & 0.2433 & 0.3241 \\
Naive RAG (Hybrid) & 0.4127 & 0.3324 & 0.3989 \\
IRCoT              & 0.4835 & 0.3751 & 0.4254 \\
MiniRAG            & 0.5429 & 0.4856 & 0.5083 \\
RAPTOR             & 0.5941 & 0.5783 & 0.5563 \\
LightRAG           & 0.7855 & 0.6756 & 0.7256 \\
HippoRAG~2         & 0.8023 & 0.7332 & 0.8032 \\
\midrule
\textbf{Ours}      & \textbf{0.8334} & \textbf{0.7554} & \textbf{0.8461} \\
\midrule
\quad w/o Auditor      & 0.8150 & 0.6823 & 0.8267 \\
\quad w/o Interpreter  & 0.6540 & 0.5434 & 0.6134 \\
\quad w/o Pathfinder\footnotemark & 0.7734 & 0.6927 & 0.7955 \\
\bottomrule
\end{tabular}
\end{table}

\footnotetext{
Instead of SRKG-based seeded retrieval and relation-aware propagation, \textit{Pathfinder} is replaced with standard dense–sparse retrieval over flat text units.
}

\subsection{Results and Analysis}
\label{sec:results}

Table~\ref{tab:main_results} reports the QA performance of all evaluated systems on the ViHERMES test set, covering answer quality ($F1$), reasoning reliability (LLM-as-a-Judge), and retrieval effectiveness ($\mathrm{Recall@}5$).
The results show a clear progression from flat, structure-agnostic RAG pipelines to multihop and graph-based approaches, indicating that Vietnamese healthcare regulatory QA requires structured multihop evidence integration.
Among naive RAG baselines, Dense retrieval improves over BM25, and Hybrid retrieval further benefits from combining semantic and lexical signals; however, these approaches remain behind multihop methods due to the lack of explicit modeling of inter-document dependencies and legal validity constraints.
IRCoT improves both $F1$ and LLM Judge scores through iterative reasoning, but remains limited by operating over unstructured text without explicit regulatory awareness.
Graph-based methods achieve markedly stronger results: MiniRAG and RAPTOR benefit from graph structure, while LightRAG and HippoRAG~2 further improve answer quality and Recall@$5$, reflecting more effective multihop evidence integration.
Overall, the proposed system achieves the best performance across all metrics, improving $F1$ by 3.1 points over HippoRAG~2 and confirming the effectiveness of structure-driven legal modeling with relation-aware propagation.

Ablation results further clarify component contributions.
Removing the \textit{Interpreter} causes the largest degradation ($F1$: 0.8334 $\rightarrow$ 0.6540), highlighting the importance of intent analysis and query routing.
Excluding the \textit{Auditor} leads to reduced reasoning reliability (LLM Judge: 0.7554 $\rightarrow$ 0.6823), underscoring the role of guardrail verification.
Replacing the \textit{Pathfinder} with flat hybrid retrieval also degrades performance, confirming the benefit of seeded retrieval with relation-aware propagation over the SRKG.

Table~\ref{tab:efficiency_results} summarizes inference efficiency and graph construction.
Compared with Naive RAG and IRCoT, graph-based approaches incur higher overhead from indexing and traversal; however, explicit legal modeling and constrained propagation achieve a favorable effectiveness--efficiency trade-off.
Despite constructing a larger SRKG, the proposed system maintains competitive inference latency (14.74s on average), comparable to RAPTOR (14.23s) and faster than HippoRAG~2.
Moreover, it outperforms LightRAG and HippoRAG~2 while using fewer graph tokens, indicating efficient context construction.

\begin{table}[t]
\centering
\caption{Inference efficiency and graph construction statistics.}
\label{tab:efficiency_results}
\begin{tabular}{lcc|ccc}
\toprule
 & \multicolumn{2}{c|}{\textbf{Inference efficiency}} 
 & \multicolumn{3}{c}{\textbf{Graph construction}} \\
\cmidrule(lr){2-3} \cmidrule(lr){4-6}
\textbf{Method}
& Avg. latency (s) 
& Avg. tokens 
& Nodes 
& Edges 
& Graph tokens \\
\midrule
Naive RAG (BM25)   & 4.1139  & 3009.8473 & --   & --    & -- \\
Naive RAG (Dense)  & 6.1139  & 3043.3433 & --   & --    & -- \\
Naive RAG (Hybrid) & 9.2348  & 3289.3246 & --   & --    & -- \\
IRCoT              & 11.8312 & 4457.4561 & --   & --    & -- \\
MiniRAG            & 13.1923 & 4224.6595 & 2988 & 4323  & 3,270,523 \\
RAPTOR             & 14.2325 & 4320.5432 & 3352 & 7234  & 3,436,236 \\
LightRAG           & 17.3236 & 4558.3425 & 3569 & 9332  & 3,684,343 \\
HippoRAG~2         & 22.0278 & 4859.3696 & 2988 & 11183 & 3,966,963 \\
\midrule
\textbf{Ours}      & 14.7415 & 4236.4620 & 3727 & 12540 & 3,397,126 \\
\bottomrule
\end{tabular}
\end{table}

\section{Conclusion}

This paper introduces ViHERMES, a benchmark for multihop QA over Vietnamese healthcare regulations that captures the legally interdependent and frequently amended nature of regulatory texts. ViHERMES provides evidence-grounded question--answer pairs with explicit reasoning annotations across diverse dependency patterns, enabling systematic evaluation of multihop regulatory QA in a low-resource language setting. To support this benchmark, we propose a SRKG and a graph-aware, multi-agent QA system that combines seeded retrieval, relation-aware propagation, and verification, achieving consistent gains over strong RAG and graph-based baselines while maintaining competitive inference latency and efficient context construction. Beyond healthcare, the proposed structure-driven formulation is applicable to other regulatory domains with hierarchical organization and formal cross-document dependencies, and future work will focus on constraining propagation volume and enriching temporal validity for improved robustness in practical deployment.

\bibliographystyle{splncs04}
\bibliography{references}

\end{document}